\documentclass[conference]{IEEEtran}
\IEEEoverridecommandlockouts
\usepackage{cite}
\usepackage{amsmath,amssymb,amsfonts}
\usepackage{graphicx}
\usepackage{textcomp}
\usepackage{xcolor}
\usepackage{algorithm}
\usepackage{algpseudocode}
\usepackage{amsmath}
\usepackage{graphics}
\usepackage{epsfig}
\usepackage{booktabs}
\def\BibTeX{{\rm B\kern-.05em{\sc i\kern-.025em b}\kern-.08em
    T\kern-.1667em\lower.7ex\hbox{E}\kern-.125emX}}
\begin{document}

\title{Really should we pruning after model be totally trained? Pruning based on a small amount of training*\\
{\footnotesize \textsuperscript{*}Note: Sub-titles are not captured in Xplore and
should not be used}
\thanks{}
}

\author{\IEEEauthorblockN{1\textsuperscript{st} Yue Li}
\IEEEauthorblockA{\textit{Nanjing University} \\
Nanjing, China \\
MF1633021@smail.nju.edu.cn}
\and
\IEEEauthorblockN{2\textsuperscript{nd} Weibin Zhao}
\IEEEauthorblockA{\textit{Nanjing University} \\
Nanjing, China \\
njzhaowb@gmail.com}
\and
\IEEEauthorblockN{3\textsuperscript{rd} Lin Shang}
\IEEEauthorblockA{\textit{Nanjing University} \\
Nanjing, China \\
shanglin@nju.edu.cn}
}

\maketitle

\begin{abstract}
Pre-training of models in pruning algorithms plays an important role in pruning decision-making. We find that excessive pre-training is not necessary for pruning algorithms. According to this idea, we propose a pruning algorithm---\textbf{Incremental pruning based on less training (IPLT)}. Compared with the traditional pruning algorithm based on a large number of pre-training, \textbf{IPLT} has competitive compression effect than the traditional pruning algorithm under the same simple pruning strategy. On the premise of ensuring accuracy, \textbf{IPLT} can achieve 8x-9x compression for VGG-19 on CIFAR-10 and only needs to pre-train few epochs. For VGG-19 on CIFAR-10, we can not only achieve 10 times test acceleration, but also about 10 times training acceleration. At present, the research mainly focuses on the compression and acceleration in the application stage of the model, while the compression and acceleration in the training stage are few. We newly proposed a pruning algorithm that can compress and accelerate in the training stage. It is novel  to consider the amount of pre-training required by pruning algorithm. Our results have implications: Too much pre-training may be not necessary for pruning algorithms.
\end{abstract}


\section{Introduction}
Deep neural networks have achieved excellent results in many competitions. The outstanding performance of the deep learning model has attracted the attention of academic and industrial circles. From AlexNet\cite{b1} to VGG-16\cite{b2} and InceptionNet\cite{b4}, it's not hard to see that the superior performance of deep learning models often depends on deeper, wider structures. A deeper deep learning model leads to better recognition accuracy, but it also creates many new problems, such as the degradation problem or the problem that the model occupies too much storage space and consumes a lot of computing resources. In order to deal with the degradation problem, \cite{b3} proposed the
Residual network model; in order to cope with the storage space required by the model, the resource consumed during the operation is too large, the researchers proposed a series of model compression and acceleration algorithms.

Most previous works on accelerating CNNs can be roughly divided into three categories, namely, matrix decomposition, quantization and pruning. The matrix decomposition of CNNs is approximately equal to decompose tensor into the product of two low-rank matrices \cite{b5}, \cite{b6}.  In the quantization of CNNs, we often change the parameters of the model from floating point number to low-bits number. For example, in \cite{b7}, the author tried to train a model with $0, 1, -1$ as its parameters. In \cite{b8} , the author not only replaces the parameters of the model with powers of 2 but also ensures that the accuracy of the network does not decrease. By converting model parameters into low bit form, floating-point operations can be converted into bit operations on specialized hardware. The pruning algorithm \cite{b9}, \cite{b11}, \cite{b27}, \cite{b28}, \cite{b30} for CNN is to prune the unimportant parameters and filters in the model by some criteria.

\begin{figure}
\centering
\includegraphics[width=.5\textwidth]{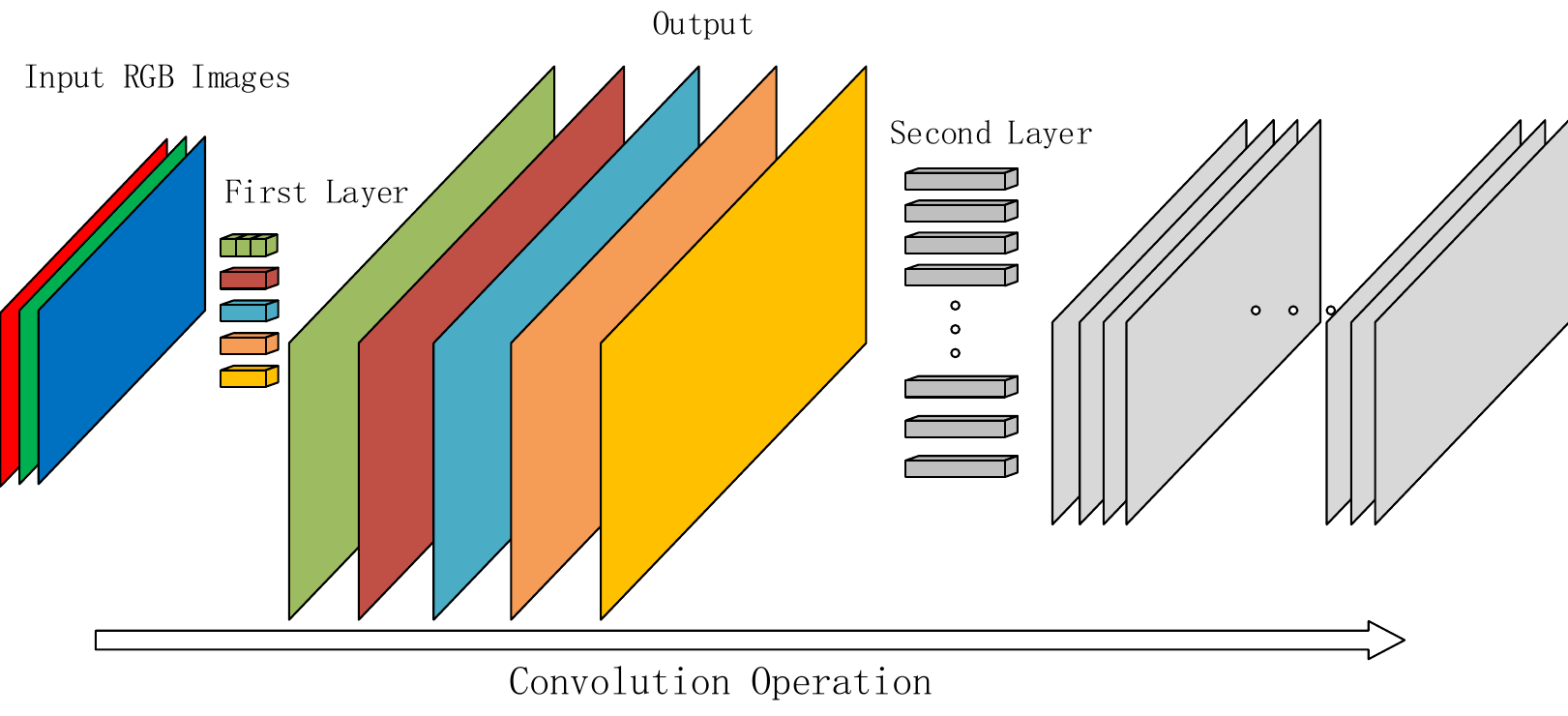}
\caption{This picture shows the convolution process of two convolution layers. The convolution network's input is an RGB image. In this picture, each cube in the graph corresponds to a filter. There are five filters in the first layer, so there are five output feature maps in the first layer. We use the same color to show the corresponding relationship between filters and output feature graphs. Notice the first filter in the first layer, which consists of three blocks. In convolution operation, each block corresponds to one input feature map.}
\label{fig:convolution_operation}
\end{figure}

In all deep learning model compression and acceleration algorithms, pruning algorithm is a kind of historical and vital algorithm. As early as 1990, \cite{b23} began to pruning the neural network. Before 2016, the pruning algorithm mainly focused on pruning the parameters of the models, distinguishing important and unimportant parameters in the model by specific criteria, cutting off unimportant parameters, and constructing a sparse convolution layer.

As in \cite{b9}, the author determines the importance of the parameter according to the size of the parameter. The larger the absolute value of the parameter, the more important the parameter is, and the less important parameter is cut off first. The algorithm mentioned in this paper is also used by \cite{b10}, which greatly compresses the deep learning model.

Then in 2016 and 2017 years, a large number of papers began to focus on the pruning of the deep learning model filter. Such as \cite{b19}, \cite{b20}, \cite{b21}, \cite{b22}, these papers began to try to slim down the structure of the model. By pruning the filters of the model, the model can be accelerated without relying on specific libraries. We can see the details of the convolution operation and the role of filters in the convolution operation through Fig.\ref{fig:convolution_operation}.

Researchers are constantly putting forward new pruning standards and strategies. In addition to improving the performances of the algorithms, in the ICLR 2019 \cite{b31}, the author puts forward some interesting ideas, which were verified by experiments. In \cite{b31}, the parameters of the pruned model were randomly initialized and trained, and the results of parameters obtained by pruning without losing or retaining were obtained. The authors suggest that the biggest significance of the pruning algorithm is not the numerical value of the parameters retained after pruning, but the refined model structure found after pruning. The author points out that: the pruned architecture itself, rather than a set of inherited “important” weights, is what leads to the efficiency benefit in the final Model. In \cite{b31}, the author believes that the meaning of pruning algorithm is searching efficient architectures.

\begin{figure}
\centering
\includegraphics[width=.5\textwidth]{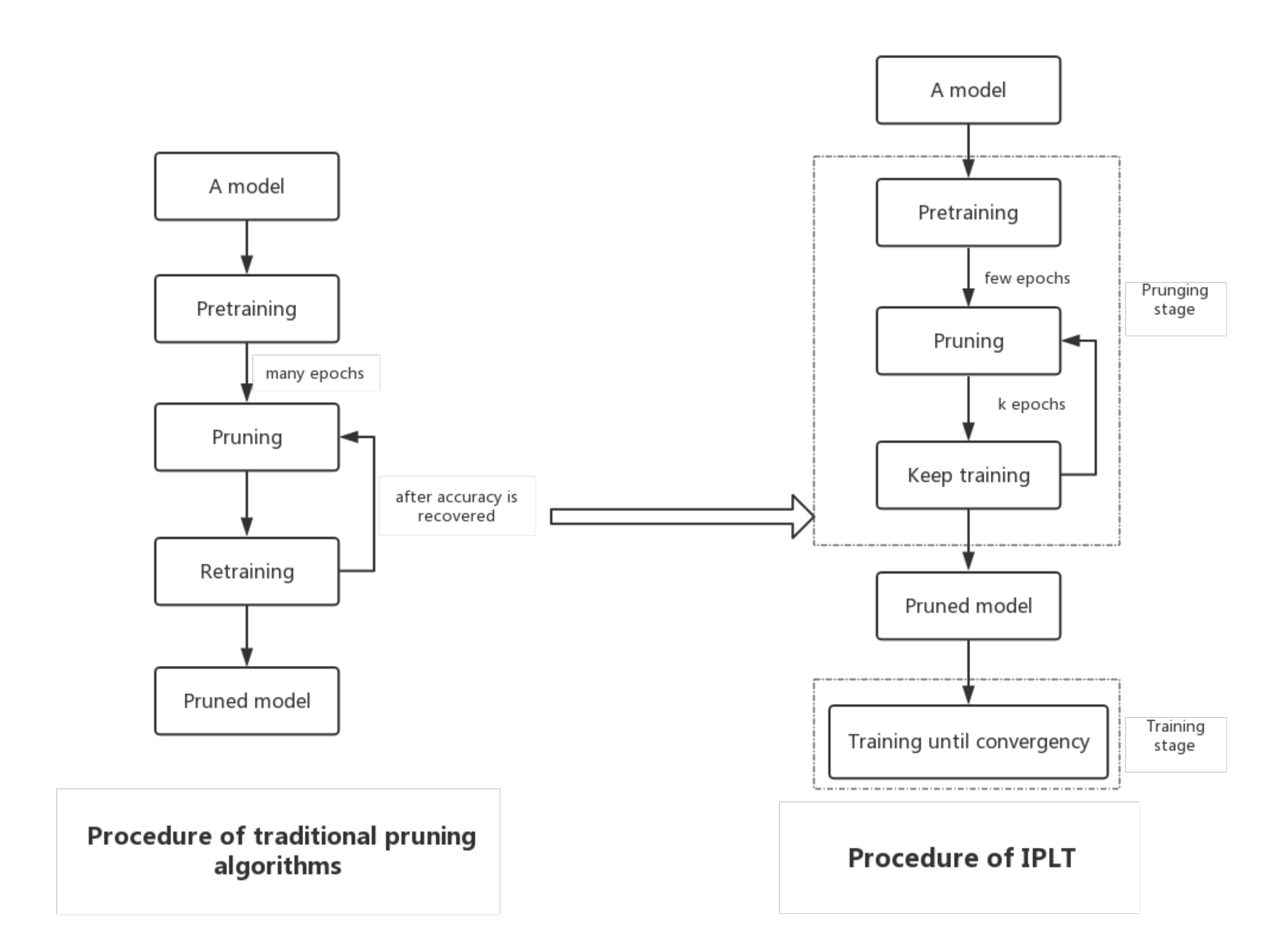}
\caption{In this picture, the general flow of the traditional pruning algorithm is on the left, and the general flow of \textbf{IPLT} is on the right. It is not difficult to find that the biggest feature of \textbf{IPLT} is: first pruning, then training. \textbf{IPLT} only trains few epochs to prune the model, and in the training phase of the model, a lot of computing resources will be reduced.}
\label{fig:pruning_contrast}
\end{figure}

Regardless of whether the parameters or the filters are pruned, almost all pruning algorithms have a general flow of pre-training models, pruning and retraining. In a word, the traditional pruning algorithm attaches great importance to the training process, especially in the pre-training stage, which usually train the model to convergence. Of course, we hope that the pruned model based on the pre-training model can be used once, but the simple parameters or filter selection criteria can not give a very effective one-step pruning decision. Therefore, a large number of researchers \cite{b11}, \cite{b14}, \cite{b20} adopt the strategy of iterative pruning, through the process of pruning and retraining on pre-trained models iteratively as shown in the left of the Fig.\ref{fig:pruning_contrast}. At the same time, some researchers \cite{b33}, \cite{b34}, \cite{b28} try to use RNN, LSTM and other models to learn the hierarchical characteristics of a network and generate a pruning decision according to the changes of training parameters. However, in the training stage of the model, such an algorithm not only needs training model, but also needs training RNN and LSTM, which makes the whole pruning and training process consume more computing resources.

$\textbf{Our Point of View:}$
The goal of pruning algorithm is to compress and speed up models more on the premise of ensuring accuracy. In my opinion, all pruning algorithms can be divided into two categories: First, these pruning algorithms \cite{b28} \cite{b35} are based on pre-trained models. These algorithms prune the models by some criterion or the forecasting model obtained by training. Therefore, when we want to prune a model with different compression ratio, we needn't to train the model from scratch. Second, this kind of algorithms \cite{b27} search effective architecture of models during the pre-training. Therefore, when we want to prune a model by second kind of algorithms with different compression ratio, we need to train the model again.

No matter the first or the second kind of algorithms, their pruning decisions are based on a large number of training epochs. In response to this situation, we raised a question: the longer the training, the more effective the pruning decision will be made?

From the Table.\ref{analysis2}, we can find that pruning decisions based on a large number of pre-training may not be more effective. Therefore, we propose the idea of pruning based on less training.

$\textbf{The Origin of \textbf{IPLT}:}$
We think that the traditional pruning algorithm pays too much attention to the pre-training process. Previous researchers often believed that the longer the pre-training process was, the more effective the pruning strategy was. Of course, if we were allowed to prune the model only once, it would be wise to do more pre-training. But in the actual pruning process, we focus on the model obtained by pruning, and do not limit the number of pruning.  And in many papers, researchers have found that the general iteratively pruning strategy is more effective, that is, the small-step pruning strategy can effectively avoid pruning excessive damage to the effective structure of the model. Therefore, there is a question worth thinking about: if we only prune a small part of the parameters or filters at a time, do we need a lot of pre-training or even training the model to convergence before the first pruning operation? We believe that only a small amount of training is needed to complete a small-step pruning on deep learning model. Based on this idea, we propose \textbf{Incremental Pruning Based on Less Training(IPLT)}.

$\textbf{Rough Flow of \textbf{IPLT}}$
The comparison between \textbf{IPLT} and traditional pruning ideas is placed on Fig.\ref{fig:pruning_contrast}. The biggest difference between \textbf{IPLT} and traditional pruning is that the model is pruned first and then trained to convergence. As shown in the figure, on the right, \textbf{IPLT} is roughly divided into pruning and training stages.\\
\textbf{Pruning Stage} \quad we need to set the hyperparameter $k$. Every time we train $k$ epoches continuously, we prune the network once. Assuming $k=5$, our ultimate goal is to prune $90\%$ of a layer of network. The pruning ratio list is $[10\%,20\%,30\%, \dots,90\%]$. As shown in Fig.\ref{fig:incremental_pruning}, we can prune $10\%$ of all filters in network's convolution layer for the first time. At the second pruning, another $10\%$ filters are pruned, the percentage of pruning reaches $20\%$. Every pruning operation, $10\%$ extra filters are pruned until $90\%$ of filters are pruned. \\
\textbf{Training Stage} \quad When pruning reaches the target pruning ratio, we stop pruning, but keep training the network until convergence.
By pruning the model step by step, our method achieves the ideal pruning ratio, and avoids the excessive pruning of the model at one time, which affects the performance.

$\textbf{The Uniqueness and Contribution of Our Work:}$\\
1.  Unlike the existing pruning algorithms, which are based on the idea of large amount of training (pre-training), we put forward the thought of pruning based on less training. According to this thought, we propose \textbf{IPLT}. The experimental results of the \textbf{IPLT} in Table.\ref{IPLTvsTrditional} prove that the model's pruning based on less training is feasible. The viewpoint of pruning based on less training is a useful supplement to the traditional pruning algorithm;\\
2. \textbf{IPLT} is the first algorithm which attempts to prune in the pre-training stage (in the first dozens of training epochs);\\
3. As far as we know, \textbf{IPLT} is the first pruning algorithm to optimize the computational complexity of model's training. \textbf{IPLT} has pruned the model in pruning stage, and will train a pruned model in training stage. Therefore, \textbf{IPLT} consumes much less computation resources.\\

\section{Related Works}
Our algorithm is about pruning. At present, pruning algorithms can be divided into two categories: weight pruning and filters pruning. We classify pruning related content as follows:
\subsection{Weights Pruning}
Many researchers try to construct sparse convolution kernels by pruning the weight of the network, so as to optimize the storage space occupied by the model. As early as around 1990, both \cite{b15} and \cite{b23} pruned the network parameters based on the second-order derivative, but this method has a high computational complexity.  In \cite{b12}, \cite{b16}, the author regularize neural network parameters by group Lasso penalty leading to sparsity on a group level. In \cite{b9}, the author judges the importance of parameters according to their value, and then prune the unimportant parameters. The \cite{b10} combine the methods in \cite{b9} with quantization, Huffman encoding, and achieve maximum compression of CNNs. \cite{b14} regularize neural network parameters by
group Lasso penalty leading to sparsity on a group level. In order to prevent overpruning, \cite{b11} proposed a parameter recovery mechanism.
By pruning the parameters, a sparse model can be constructed. This kind of method can compress the model storage. Because the application of these pruned models always depend on specific libraries, computational optimization is not sufficient. So in the past two years, many researchers have turned their attention to pruning filters.

\subsection{Filters Pruning}
In the past two years, there has been a lot of work about filters pruning algorithms. Most papers use certain criteria to evaluate filters, and ultimately prune unimportant filters. In 2017, \cite{b17}  try to use $l1-norm$ to select unimportant filters. \cite{b18} uses the scaling factor $\gamma$ in batch normalization as an important factor, that is, the smaller the $\gamma$ is, the less important the corresponding channel is, so that filters can be pruned. \cite{b21} proposes a Taylor expansion based pruning criterion to approximate the change in the cost function induced by pruning.
In addition to pruning filters through specific criteria, some researchers also proposed new ideas. \cite{b28} proposed utilizing a long short-term memory (LSTM) to learn the hierarchical characteristics of a network and generate a pruning decision for each layer. \cite{b29} proposed a model pruning technique that focuses on simplifying the computation graph of a deep convolutional neural network. In \cite{b27}, the author  proposed a Soft Filter Pruning (SFP) method to accelerate the inference procedure of deep Convolutional Neural Networks.

In addition to the above papers, some researchers \cite{b12}, \cite{b13} have proposed algorithms that can be used to prune both parameters and filters.

\subsection{Discussion}
Compared with the second kind of pruning algorithms, \textbf{IPLT}'s feature is the optimization of computational complexity in training stage. \textbf{IPLT} uses few epochs' training to find filters which are not very important in the model, and then prune them. This avoids consuming a lot of computing resources in the model's training stage.

We need to emphasize the difference between our incremental pruning and traditional iteration pruning. Some traditional pruning algorithms also gradually increase the pruning ratio when pruning the model. This kind of pruning method is often called iterative pruning. Our algorithm and this kind of algorithm have chosen to increase the pruning ratio step by step in pruning the model, the difference is that our method does not rely on convergent network, that is, we will not wait for the CNNs model to be trained to converge before the next pruning. This saves a lot of computing resources and makes the method simpler and easier.

There is a work \cite{b27} trying to combine training with pruning. In this paper, models are pruned in a soft mode. The biggest difference between us is that we actually prune some filters of models, and \cite{b27} only add a mask to the parameters or filters, temporarily excluding the parameters or filters from the forward operation, in fact, these parameters are still updated. \cite{b27} belongs to the second kind of pruning algorithms. \textbf{IPLT} can save the computational complexity in training stage， but \cite{b27} can't. Apparently, \textbf{IPLT} is different from \cite{b27}. But in a way, we think \textbf{IPLT} is \cite{b27} combined with the thought of pruning based on less training.

\section{Methodology}
\subsection{Preliminaries}\label{AA}
In this section, we will formally introduce the symbol and annotations. We use $\left\{\mathcal{W}_i \in \mathbb{R}^{O_i \times I_i \times K \times K} , 1\leq i \leq L \right\}$ and $\left\{b_i \in \mathbb{R}^{O_i} , 1\leq i \leq L \right\}$to denote ith convolutional layer's weights and bias, L is the number of layers. $\textbf{I}_i$ and $\textbf{O}_i$ denote the number of input and output feature maps in ith layer, so the input tensor of $i-th$ layer can be represented by $\textbf{FI}_i$ and its size is $I_i \times H_i \times \mathcal{W}_i$, output tensor of $i-th$ layer can be represented by $\textbf{FO}_i$ and its size is $O_i \times H_{i+1} \times \mathcal{W}_{i+1}$. Obviously, $\textbf{FI}_(i+1) = \textbf{FO}_i$, and in CNNs with RGB images $\textbf{I}_1 = 3$. $\mathcal{F}_{i,j}$ represent $j-th$ filter in CNNs' $i-th$ layer, $\mathcal{F}_{i,j} \in \mathbb{R}^{I_i \times K \times K}$. The convolutional operation of $i-th$ layer can be written as:

\begin{equation}
FO_{i} = \mathcal{W}_i \ast FI_i + b_i, 1\leq i \leq L
\label{equ.1}
\end{equation}

Equation.\ref{equ.2} can be seen as a composite of  $O_i$ filters' convolutional operation:

\begin{equation}
FO_{i,j} = \mathcal{F}_{i,j} \ast FI_i, 1\leq j \leq O_i
\label{equ.2}
\end{equation}

where $FO_{i,j}$ is the $jth$ output feature map of $i-th$ layer. Obviously, $FO_{i} = \left\{FO_{i,j},  1\leq j \leq O_i \right\}$.

We use $R_i$ to indicate the percentage of filters pruned from $i-th$ Layer to all filters in the same layer. In this case, the number of filters and output feature maps in $i-th$ layer will be reduced to $(1-R_i)O_i$, so the parameters of $i-th$ layer will be reduced from $O_i \times I_i \times K \times K$ to $(1-R_i)O_i \times I_i \times K \times K$. Not only the $i-th$ layer's filters pruned but also the $(i+1)-th$ layer will be affected. As shown in \ref{fig:select_filters_to_prune}, the filters in the $i-th$ layer are pruned, so in the next layer, each filter should be slimmed down. $\mathcal{F}_{i+1,j} \in \mathbb{R}^{O_i \times K \times K}$ should be transformed into $\mathcal{F}^{'}_{i+1,j} \in \mathbb{R}^{(1-R_i)O_i \times K \times K}$

\subsection{How to Select Filters}
\begin{figure}
\centering
\includegraphics[width=.5\textwidth]{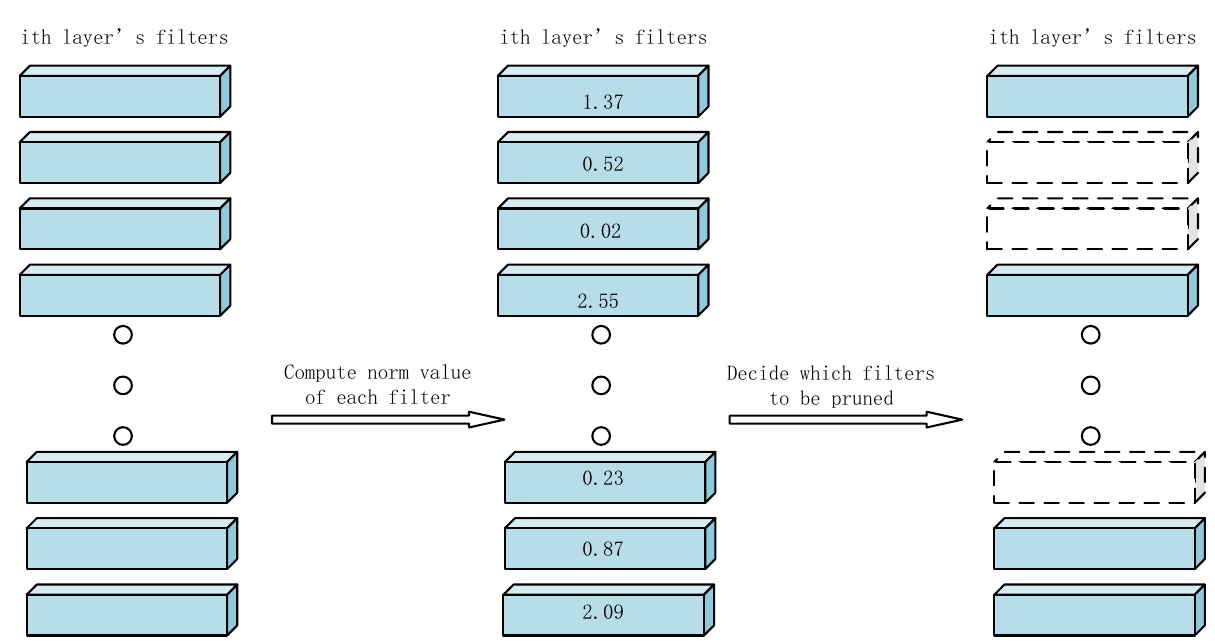}
\caption{This picture shows a pruning operation. We calculate the $\emph{l}_2-$norm of filters in $i-th$ layer, and prune filters with smaller norm by sorting. The first column is all filters in $i-th$ layer. The second column shows that we have calculated the norm values corresponding to each filter in $i-th$ layer. The filters drawn by dashed lines in the third column are pruned. Obviously, filters with smaller norm values are more pruned first. This figure only shows the importance comparison of filters in intra-layer mode; if it is global comparison mode, then we will sort the norm values of filters in all convolution layers and prune filters with small norm values. }
\label{fig:select_filters_to_prune}
\end{figure}

We chose $\emph{l}_2-$norm to measure the importance of each filter as Eq.\ref{norm}.
\begin{equation}
    \|\mathcal{F}_{i,j}\|_p = \sqrt[p]{\sum_{n=1}^{I_i} \sum_{k_1=1}^K \sum_{k_2=1}^K |\mathcal{F}_{i,j}(n,k_1,k_2)|^p}
\label{norm}
\end{equation}
In general, filters with smaller $\emph{l}_2-$norm result in relatively small activation values, so we think these filters are even less important to the model. The filters with smaller $\emph{l}_2-$norm will be pruned firstly. When we compare the $\emph{l}_2-$norm of filters, we can choose either intra-layer comparison (intra-layer mode) or full-network comparison (global mode). The only difference between intra-layer comparison and full-network comparison is whether all filters in the whole network are sorted together (full-network comparison) or within each layer (intra-layer comparison) when filters are sorted in norm value. We show the general procedure in Alg.\ref{alg:global}. In Fig.\ref{fig:select_filters_to_prune}, we compare the importance of filters in each layer (intra-layer), and pruning the filters with smaller $\emph{l}_2-$norm values.

\begin{algorithm}[htb]
  \caption{globally comparision pruning}
  \label{alg:global}
  \begin{algorithmic}[1]
    \Require
      training data:X, training epoches: $epoch_{max}$,
      incremental pruning sequence: $L = \left\{R_1, R_2, ..., R_n\right\}$ ,
      model parametes: $\mathcal{W} = \left\{\mathcal{W}_i , 1\leq i \leq L \right\}$,
      A pre-trained CNN model, ${W_l,1\leq l\leq L}$;\\
    \Ensure
      Pruned models:${\hat{W_l}, 1\leq l\leq L}$;
    \State Initial: ${W_l}$, ${1\leq l\leq L}$ ; hyper-paramter: k;
    \label{code:fram:extract}
    \State for ${epoch=1, 2, \ldots, epoch_{max}}$, do;
    \label{code:fram:trainbase}
    \State \qquad if epoch\%k==0 and epoch<=k*len(L):
    \label{code:fram:add}
    \State \qquad \qquad for $i=0, 1\leq i\leq L$, do:
    \label{code:fram:classify}
    \State \qquad \qquad \qquad Calculate the $\emph{l}_2-$norm for each filter $\mathcal{F}_{i,j}, 1\leq j \leq I_i$ ;
    \label{code:fram:select} \\
    \State \qquad  Prune $R_{ind}\ast\sum_{i=1}^LO_i$ filters with minmum $\emph{l}_2-$norm value in all layers(global) or per layer(intra-layer);
    \label{code:fram:classify}
    \State \qquad Update model parameters $W$ based on $X$;
    \label{code:fram:add}
    \label{code:fram:select}
    \Return A pruned model;
  \end{algorithmic}
\end{algorithm}

Experiments show that globally sorting filter norms and pruning filters can better guarantee network performance.

\subsection{The thought of incremental pruning}
\begin{figure}
\centering
\includegraphics[width=.5\textwidth]{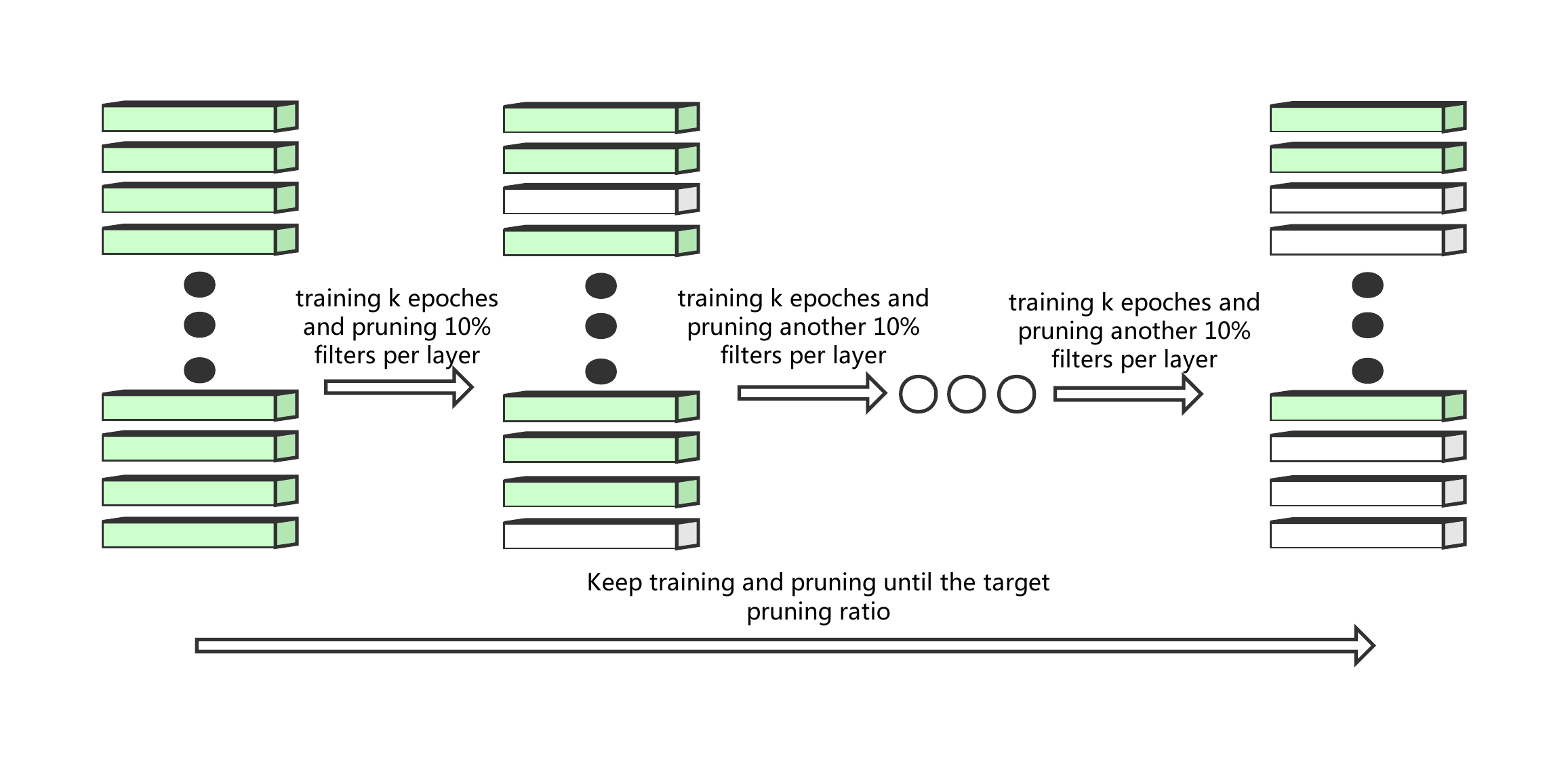}
\caption{In this picture, we show the general process of incremental pruning. In the figure, filters in a convolution layer are pruned. A column of cubes in the figure is all filters in this convolution layer. White cuboids denote filters that have been pruned. Obviously, with each step of pruning, more and more filters have been pruned.}
\label{fig:incremental_pruning}
\end{figure}

The goal of our experiment is to complete the pruning of the network model in the process of deep learning model training. Because the model is in training stage, the parameters are not stable. The importance of the parameters is determined directly based on the size of the parameters. It is difficult to get a good network structure relationship by dividing which parameters (filter) are pruned and which are retained at one time. So we don't directly pruning the network in one step, but step by step as shown in the Fig.\ref{fig:incremental_pruning}. For example, according to the sequence $[10\%, 20\%, 30\%, 40\%, 50\%, 60\%, 70\%]$, we will first cut 10\% of the filters, then cut off the extra 10\% (achieve 20\% pruning rate). By gradually pruning, we keep pruning a small number of filters which are not important, and finally achieve the desired pruning rate for our network.

\subsection{How to Prune}
\begin{figure}
\centering
\includegraphics[width=.5\textwidth]{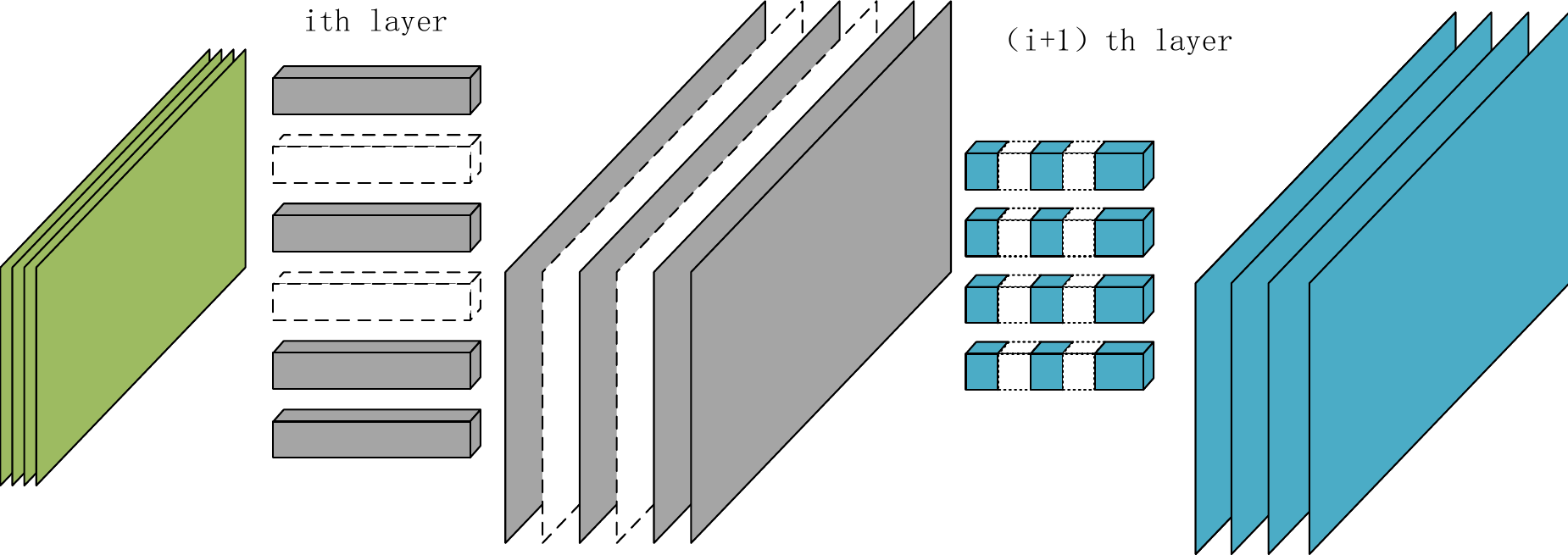}
\caption{This picture shows that when we prune filters of one convolution layer, the next layer needs to make some adjustments at the same time. In this picture, grey and blue filters correspond to two adjacent convolution layers. We pruned the filters on the $i-th$ layer. We prune the second and third filters of $i-th$ layer, so the second and fourth output feature maps of $i-th$ layer are also pruned. Obviously, the number of input feature maps in $(i+1)-th$ layer is reduced. The $(i+1)-th$ Layer does not need to convolute the second and fourth pruned feature maps, so the number of filters in $(i+1)-th$ layer does not change, and the size of each filter should be modified to $\frac{2}{3}$ of the original.}
\label{fig:filters_pruning}
\end{figure}

In Fig.\ref{fig:filters_pruning}, we show the operations performed when pruning a layer.

If we only prune the $i-th$ convolution layer of CNNs. We assume that there are $O_i$ filters in $i-th$ layer of the network. The pruning ratio of filters in this layer is $R_i$, and the size of convolution kernel is $K_i \ast K_i$. Obviously, there are $R_i \times O_i$ filters cut off in $i-th$ layer, and the number of output feature maps in $i-th$ layer will change from $O_i$ to $Ri \times O_i$. Just as the gray filters in the Fig.\ref{fig:filters_pruning} are cut off, the blue filters in the next layer need to be adjusted in size: originally, in $(i+1)th$ layer, each convolution $\mathcal{F}_{i+1,j} \in R^{O_i \star K_{i+1}\star K_{i+1}}$ , now becomes $\mathcal{F}^{'}_{i+1,j} \in R^{R_i \ast O_i \star K_{i+1}\star K_{i+1}}$. So when the pruning ratio of the $i-th$ layer is $R_i$, the parameters of $O_i \ast R_i \times Ii \times K_i \times K_i$ will be cut off in this layer, and $O_{i+1} \times O_i \ast R_i \times K_{i+1} \times K_{i+1}$ parameters should also be cut in the $(i+1)th$ layer. Because of the pruning ratio of $R_i$ in the $i-th$ layer, the pruned parameters of the whole network are $R_i* (O_i\times I_i \times K_i \times K_i+O_{i+1} \times O_i \times K_{i+1} \times K_{i+1} \times K_{i+1})$. In order to facilitate readers to better understand our network compression, we will propose two concepts of pruning rate: 1. Filters Pruning Ratio ($FPR$); 2. Parameters Pruning Ratio ($PPR$). $FPR$ represents the percentage of filters that have been pruned to total filters. PPR is the proportion of the number of parameters that are pruned to the total number of parameters. Obviously we can calculate $FPR$, $PPR$ for each layer, or the $FPR$, $PPR$ for the whole network.

Suppose we only prune the filters of the $i-th$ layer, and the ratio of pruning is $R_i$. We use $FPR_i$ to represent $FPR$ in $i-th$ layer and $FPR_{all}$ to represent $FPR$ in the whole network. Eq.\ref{con:FPR} illustrates the calculation method of $FPR_i$ and $FPR_{all}$.
\begin{equation}
FPR_i = R_i = \frac{O_i \ast R_i}{O_i}; \\
\quad FPR_{all} = \frac{O_i \ast R_i}{\sum_{i=1}^LO_i}
\label{con:FPR}
\end{equation}

Obviously, whether $FPR_{all}$ or $PPR_{all}$, the higher the numerical value, the more we prune the network, the more we compress and accelerate the model. In Table.\ref{FPR_PPR}, we mainly observed $FPR_i$ and $PPR_{all}$. Obviously, $FPR_i$ mainly reflects the pruning of filters in each layer of CNNs. $PPR_{all}$ mainly reflects the situation of model compression and acceleration.

Based on CIFAR-10 dataset, we prune on VGG-16 network. We take k = 5 and gradually increase the pruning ratio ($FPR$) of filters according to the series $[10\%,20\%,30\%, \dots,70\%]$. Here we choose the intra-layer mode to find fiters which should be pruned. Therefore, when the model has been pruned, $FPR_i$ is equal to 70\% for all convolutional layers. In Table.\ref{FPR_PPR}, we list the $FPR_{all}$, $PPR_{all}$ and $FPR_i$, $PPR_i$ of each layer of the network after pruning .

\begin{table}[htbp]
 \caption{FPR, PPR of Pruned VGG-19}
 \begin{tabular}{lll}
  \toprule
  Layer&FPR&PPR\\
  \midrule
  Conv1&3.125&3.125\\
  \hline
  Conv2&3.125&6.152\\
  \hline
  Conv3&3.125&6.152\\
  \hline
  Conv4&3.906&6.909\\
  \hline
  ...&...&...\\
  \hline
  Conv15&90.039&99.358\\
  \hline
  Conv16&80.859&98.093\\
  \hline
  Fc1&80.859&80.859\\
  \hline
  Fc2&0.0&0.0\\
  \hline
  Fc3&0.0&0.0\\
  \hline
  Total&-&-\\
  \bottomrule
 \end{tabular}
\label{FPR_PPR}
\end{table}

\section{Experiments}

\subsection{Benchmark Datasets and Experimental Setting}
\textbf{Datasets Selection:}We empirically apply our methods on two benchmark datasets:MNIST \cite{b24} ,CIFAR-10 \cite{b25}. The MNIST dataset consists of 60,000 $28\times28$ black-and-white images. The CIFAR-10 dataset consists of 60,000 $32\times32$ color images. There are 50,000 training images and 10,000 test images. The images in MNIST and CIFAR-10 datasets are both divided into 10 classes, with 6,000 images in each class.

\textbf{Two Networks:} We choose a normal CNN and VGGNet\cite{b2} to verify the effectiveness of our method. All the experiments are implemented with PyTorch on one NVIDIA GPU. For the reason of convenience of implementation, we choose to realize the pruning stage in a soft mode --- during both pruning stages of \textbf{IPLT} and traditional pruning algorithms, we add a mask which consists of 0 and 1 to realize the equivalent effect of pruning. But in \textbf{IPLT}, we prune the model after pruning stage in reality. After the pruning stage of \textbf{IPLT}, we create a new model with fewer filters is created. We copy the remaining parameters of the modified layers into the new model. Then in training stage of \textbf{IPLT}, we will train the actually pruned model.

\subsection{Contrast Experiments between intra-layer mode and global mode}
Firstly, we compare two methods of filter importance ranking (intra-layer, global mode) when pruning filters. The first experiment is based on MNIST and a normal CNN. In this experiment, the hyperparameter k is set to 2 and the pruning proportion sequence is $[10\%, 20\%, 30\%, \dots]$. This means that if the final filter pruning ratio is $70\%$, in the first 14 epochs, we prune the model once every 2 epochs.

After the first 14 epochs, we do no pruning operation and only train the network.

The second experiment is based on CIFAR 10 dataset and VGG-19 network. Our parameter settings are basically the same as the original VGG-19 in \cite{b2}. In \cite{b2}, a batch normalization layer is added after each convolution layer \cite{b32}. The hyperparameter k is set to 5. This means that if the final filters pruning ratio is 70\%, we will do the pruning of the filters once every 5 epochs in the first 35 epochs. After the first 35 epochs, we do no pruning and only train the network. The comparative results of the experiments are shown in the Table.\ref{mnist} and Table.\ref{intra_layer_global_cifar10}.

By comparing the experimental results in the two tables, we can clearly find that in larger dataset, the accuracy of the pruned network can be better guaranteed by sorting the importance of the whole network filters and pruning the less important filters (global mode).

\begin{table}[!htbp]
\caption{A CNN on MNIST}
\begin{tabular}{|l|l|l|l|}
\hline
{Mode} & {$FPR_{all}$} & $PPR_{all}$ & $Accuracy(\%)$ \\ \hline
no\_prune & 0.00 & 0.00 & 99.35 \\ \hline
$Intra\_layer$ & 60.00 & 83.05 & 99.32 \\
$Intra\_layer$ & 70.00 & 90.65 & 99.17 \\ \hline
Global & 60.00 & 84.72 & 99.3\\
Global & 65.00 & 87.89 & 99.31\\
Global & 70.00 & 95.47 & 99.35\\ \hline
\end{tabular}
\label{mnist}
\end{table}

\begin{table}[!htbp]
\caption{VGG-19 on CIFAR-10}
\begin{tabular}{|l|l|l|l|}
\hline
Mode&$FPR_{all}$&$PPR_{all}$&Accuracy(\%) \\ \hline
$no\_prune$&0.00&0.00&94.21\\ \hline
$Intra\_layer$&70.0&88.53&91.74\\
$Intra\_layer$&60&81.75&92.30\\ \hline
Global&70&85.00&94.20\\
Global&60&80.76&-\\ \hline
\end{tabular}
\label{intra_layer_global_cifar10}
\end{table}

In order to further demonstrate the performance of our algorithm, we have done some comparative experiments with other traditional pruning algorithms.

\subsection{Contrast Experiments between \textbf{IPLT} and Traditional Pruning Algorithm(First Kind)}
\subsubsection{Compared with Traditional Pruning Algorithm Implemented by Ourselves}
This experiment is based on VGG-19 network and CIFAR-10 dataset. But the VGG-19 in this paper is slightly different from the original VGG-19 \cite{b2}. In the experiment, we prune the VGG-19 on the CIFAR-10 dataset. Each convolution layer is followed by a batch normalization layer in the VGG-19 and we remove its FC layers except the last layer for classification.

The comparative results of the experiments are shown in the Table.\ref{IPLTvsTrditional}.

\begin{table}[!htbp]
\caption{In this table, we do comparative experiments based on VGG-19 and CIFAR-10 dataset. $FPR$ represents the percentage of filters that have been pruned to total filters. Accuracy is the optimal accuracy of the model on the test data set.}
\begin{tabular}{|l|l|l|}
\hline
Model&FPR(\%)&Accuracy(\%)\\ \hline
baseline&0.00&94.21\\ \hline
IPLT&65.0&94.22\\
IPLT&70.0&94.20\\
IPLT&80.0&93.20\\ \hline
Traidit&55.0&94.15\\
Traidit&60.0&93.90\\
Traidit&65.0&10.00\\
Traidit&70.0&10.00\\
Traidit&70.0&10.00\\ \hline
\end{tabular}
\label{IPLTvsTrditional}
\end{table}

In pruning, based on $\emph{l}_2-$norm, the standard of gobal mode pruning is very simple, so when pruning proportion is too high, pruning algorithm often prunes a layer completely. This is the reason why there are some 10\% in Table.\ref{IPLTvsTrditional}. Apparently, \textbf{IPLT} can achieve a higher $FPR$ based on the same pruning criteria. Baseline is based on the results of the complete, no pruned VGG-19.\\
\indent As shown in Table.\ref{IPLTvsTrditional}, when compressing the model with same ratio, \textbf{IPLT} can maintain better accuracy. In addition to comparing with the traditional pruning algorithm implemented by ourselves, we also compared \textbf{IPLT} with \cite{b35}.

\subsubsection{Contrast Experiments between \textbf{IPLT} and 'Efficient Pruning'\cite{b35}}

To further prove the effectiveness of \textbf{IPLT} and the thought of pruning based on less training(pre-training). We did contrast experiments between \textbf{IPLT} and \cite{b35}. In \cite{b35}, the author proposes a pruning algorithm which is also based on norm values. They consider the independent and greedy strategies for filters selection which are similar to strategies in \textbf{IPLT}.

To ensure the objectivity of our baseline, we adopt the same dataset CIFAR-10 and same model structure as \cite{b35}. Based on Pytorch library, we use 'torch.save(model.state\_dict())' to save the the model before and after pruning.

\begin{table*}[htbp]
 \caption{VGG-16 on CIFAR-10. efficient's baseline is the baseline in \cite{b35}. This paper uses the same model structure and data set as \cite{b35}. The 'efficient's result' refers to the result of \cite{b28}}
 \begin{tabular}{llllll}
  \toprule
  Model&$FPR$&Model Size(MB)&FLOPs&Pruned FLOPs(\%)&Accuracy(\%)\\
  \midrule
  \hline
  efficient's baseline\-&-&-&$3.13 \times 10^8$&0.00&93.25\\
  \hline
  efficient's result\-&-&-&$2.06 \times 10^8$&34.20&93.40\\
  \hline
  ours\_baseline&0.00&58.9&$3.13 \times 10^8$&0.00&94.33\\
  \hline
  \textbf{IPLT}&60&9.7&$1.52 \times 10^8$&51.36&94.35\\
  \textbf{IPLT}&70&6.0&$1.27 \times 10^8$&59.54&94.05\\
  \bottomrule
 \end{tabular}
\label{analysis2}
\end{table*}

We propose the \textbf{IPLT} according to the thought of pruning based on less training. As shown in Table.\ref{analysis2}, based on similar pruning strategy, \textbf{IPLT} has better performance than \cite{b35}. We think that this is a better illustration of the effectiveness of \textbf{IPLT}. This is also a good illustration of the thought that too much pre-training may be unnecessary for pruning algorithm.

\subsection{Ours vs 'Where to Prune'\cite{b28}}
We analyze the compression effect and acceleration effect of \textbf{IPLT} algorithm. The dataset is CIFAR-10 and the model is VGG-19. To ensure the objectivity of our baseline, we adopt the same model structure as \cite{b28}. The analysis results are showing in following Table.\ref{analysis}.
Based on Pytorch library, we use 'torch.save(model.state\_dict())' to save the the model before and after pruning.

\begin{table*}[htbp]
 \caption{VGG-19 on CIFAR-10. Where's baseline is the baseline in \cite{b28}. This paper uses the same model structure and data set as \cite{b28}. The 'where's result' refers to the result of \cite{b28}}
 \begin{tabular}{llllll}
  \toprule
  Model&$FPR$&Model Size(MB)&FLOPs&Pruned FLOPs(\%)&Accuracy(\%)\\
  \midrule
  \hline
  where's baseline\-&-&-&$3.9 \times 10^8$&0.00&93.66\\
  \hline
  where's result\-&-&-&$5.98 \times 10^7$&84.70&93.30\\
  \hline
  ours\_baseline&0.00&70.9&$3.9 \times 10^8$&0.00&94.21\\
  \hline
  \textbf{IPLT}&65&9.7&$1.79 \times 10^8$&53.92&94.22\\
  \textbf{IPLT}&70&8.6&$1.64 \times 10^8$&57.76&94.20\\
  \textbf{IPLT}&75&6.3&$1.41 \times 10^8$&63.70&94.12\\
  \textbf{IPLT}&80&4.8&$1.18 \times 10^8$&69.45&94.02\\
  \bottomrule
 \end{tabular}
\label{analysis}
\end{table*}

We must admit that in terms of model acceleration, \textbf{IPLT} is not as good as \cite{b28}. However, in \cite{b28}, the author uses a LSTM to learn the hierarchical characteristics of a network. This greatly increases the computational cost of the pre-training stage. Such pruning decision-making method is more complex, while \textbf{IPLT} only uses relatively simple pruning criteria and strategy. Therefore, it is acceptable that the model acceleration effect of IPLT is slightly weak. And as shown in the results, the \cite{b28} method has a greater loss of pruning accuracy, while our method has a smaller loss of precision.
As shown in Table.\ref{analysis}, the speed of VGG-19 pruned by \textbf{IPLT} is about 2.5x faster with almost no accuracy loss. Because \textbf{IPLT} starts pruning after a small amount of training on the model, it can also achieve about 2.5x acceleration in the training stage. There is no doubt that the computational resource consumption during the model's training stage has also been greatly optimized.

\subsection{Conclusion}
In this paper, we propose the thought of pruning based on less training. We combine this thought with thought of incremental and propose \textbf{IPLT}. \textbf{IPLT} also can be seen as a combination of \cite{b27} (second kind of pruning algorithms) and the thought of pruning based on less training. As shown in Table.\ref{analysis2}, Based on simple pruning criteria, \textbf{IPLT} can compress and speed up the VGG-16 more than traditional pruning algorithm \cite{b35}(first kind of pruning algorithms). This undoubtedly proves that sometimes pruning algorithms based on less training is more effective than pruning algorithms based on a large amount of training(pre-training).

We think that the thought of pruning based less training is not only novel, but also necessary.

Novelty: Nobody has ever questioned the necessity of so much pre-training. Therefore, no one has considered whether pruning based on more training epochs(pre-trained models) can guarantee more model compression and acceleration. This is the first paper proposing these two questions. We find that at least sometimes the pruning algorithms based on less training(pre-training) will compress and speed up models more.

Necessity: 1. From the Table.\ref{analysis2}, the contrast experiments between \textbf{IPLT} and \cite{b27} prove that the thought of pruning based on less training can be combined with pruning criteria and pruning decision-making methods in the first kind of pruning algorithms, so that we may compress and speed up models more; 2. \textbf{IPLT} can be seen as a combination of \cite{b27} and the thought of pruning based on less training. \textbf{IPLT} can prune models in the training stage and optimize the computational complexity in training stage. This shows that the thought of pruning based on less training can be combined with the second kind of pruning algorithms, so that we can optimize the computational complexity of these algorithms in the training stage.

\vspace{12pt}
\color{red}

\end{document}